\documentclass[10pt,journal,compsoc]{IEEEtran}

\ifCLASSOPTIONcompsoc
  \usepackage[nocompress]{cite}
\else
  \usepackage{cite}
\fi

\ifCLASSINFOpdf
   \usepackage[pdftex]{graphicx}
\else
\fi

\usepackage{amsmath}
\usepackage{color}
\usepackage{amssymb}
\usepackage{multirow}
\usepackage{subcaption}
\usepackage{siunitx}
\usepackage{xfrac}
\usepackage{url}

\newcommand{\rad}[1]{\SI{#1}{\radian}}
\newcommand{\rpm}{\sbox0{$1$}\sbox2{$\scriptstyle\pm$}
  \raise\dimexpr(\ht0-\ht2)/2\relax\box2 }
\begin{document}
\title{Light Field Super-Resolution using a Low-Rank Prior and Deep Convolutional Neural Networks}

\author{Reuben~A.~Farrugia,~\IEEEmembership{Senior Member,~IEEE,}
        Christine Guillemot,~\IEEEmembership{Fellow,~IEEE,}

\IEEEcompsocitemizethanks{\IEEEcompsocthanksitem R.A. Farrugia is with the Department of Communications and Computer Engineering, University of Malta, Malta,
e-mail: (reuben.farrugia@um.edu.mt).\protect\\
E-mail: see http://www.michaelshell.org/contact.html
\IEEEcompsocthanksitem C. Guillemot is with the Institut National de Recherche en Informatique et en Automatique, Rennes 35042, France,
e-mail: (christine.guillemot@intria.fr).}
\thanks{Manuscript received January, 2018; revised xxx.}}


\IEEEtitleabstractindextext{%
\begin{abstract}
Light field imaging has recently known a regain of interest due to the availability of practical light field capturing systems that offer a wide range of applications in the field of computer vision. However, capturing high-resolution light fields remains technologically challenging since the increase in angular resolution is often accompanied by a significant reduction in spatial resolution. This paper describes a learning-based spatial light field super-resolution method that allows the restoration of the entire light field with consistency across all sub-aperture images. The algorithm first uses optical flow to align the light field and then reduces its angular dimension using low-rank approximation. We then consider the linearly independent columns of the resulting low-rank model as an embedding, which is restored using a deep convolutional neural network (DCNN). The super-resolved embedding is then used to reconstruct the remaining sub-aperture images. The original disparities are restored using inverse warping where missing pixels are approximated using a novel light field inpainting algorithm. Experimental results show that the proposed method outperforms existing light field super-resolution algorithms, achieving PSNR gains of 0.23 dB over the second best performing method. This performance can be further improved using iterative back-projection as a post-processing step.
\end{abstract}

}

\maketitle

\IEEEdisplaynontitleabstractindextext

\IEEEpeerreviewmaketitle

\IEEEraisesectionheading{\section{Introduction}\label{sec:introduction}}

\IEEEPARstart{L}{ight}
field imaging has emerged as a promising technology for a variety of applications going from photo-realistic image-based rendering to computer vision applications such as 3D modeling, object detection, classification and recognition.
As opposed to traditional photography which captures a 2D projection of the light in the scene, light fields collect the radiance of light rays along different directions \cite{Ng2005,Wilburn2005}.
This rich visual description of the scene offers powerful capabilities for scene understanding and for improving the performance of traditional computer vision problems such as depth sensing, post-capture refocusing, segmentation, video stabilization and material classification to mention a few. 

Light fields acquisition devices have been recently designed, going from rigs of cameras \cite{Wilburn2005} capturing the scene from slightly different viewpoints to plenoptic cameras using micro-lens arrays placed in front of the photo-sensor \cite{Ng2005}. These acquisition devices offer different trade-offs between angular and spatial resolution. Rigs of cameras capture views with a high spatial resolution but in general with a limited angular sampling hence large disparities between views. On the other hand, plenoptic cameras capture views with a high angular sampling, but at the expense of a limited spatial resolution. In plenoptic cameras, the angular sampling is related to the number of sensor pixels located behind each microlens, while the spatial sampling is related to the number of microlenses. 

Light fields represent very large volumes of high dimensional data bringing new challenges in terms of capture, compression, editing and display. 
The design of efficient light field image processing algorithms, going from analysis, compression to super-resolution and editing has thus recently attracted interest from the research community. A comprehensive overview of light field image processing techniques can be found in \cite{wu_2017}.

This paper addresses the problem of light field spatial super-resolution.
Single image super-resolution has been an active field of research in the past years, leading to quite mature solutions. However, super-resolving each view separately using state of the art single-image super-resolution techniques would not take advantage of light field properties, in particular of angular redundancy which depends on scene geometry\cite{Liang2015}. Moreover, considering each sub-aperture image as a separate entity may reconstruct light fields which are angularly incoherent \cite{Farrugia2017}. 

Assuming that the low-resolution light field captures different views of the same scene taken with sub-pixel misalignment, the problem can be posed as the one of recovering the high-resolution (HR) views from multiple low- resolution images with unknown non integer translational misalignment. A number of methods hence proceed in two steps. A first step consists in estimating the translational misalignment using depth or disparity estimation techniques. The HR light field views are then found using Bayesian or variational optimization frameworks with different priors.
This is the case in \cite{Levin_2008} and \cite{Bishop2012} where the authors first recover a depth map and formulate the spatial light field super-resolution problem either as a simple linear problem \cite{Levin_2008} or as a Bayesian inference problem \cite{Bishop2012} assuming an image formation model with Lambertian reflectance priors and a depth-dependent blurring kernel. A Gaussian mixture model (GMM) is proposed instead in \cite{Mitra2012} to address denoising, spatial and angular super-resolution of light fields. The reconstructed 4D-patches are estimated using a linear minimum mean square error (LMMSE) estimator, assuming a disparity-dependent GMM for
the patch structure. In \cite{Wanner2014}, the geometry is estimated by computing structure tensors in the Epipolar Plane Images (EPI). A variational optimization framework is then used to spatially super-resolve the different views given their estimated depth maps and to increase the angular resolution.
 
Another category of methods is based on machine learning techniques which learn a model of correspondences between low- and high-resolution data. 
In \cite{Farrugia2017}, the authors learn projections between 
low-dimensional subspaces of 3D patch-volumes of low- and high-resolution, using ridge regression. 
Data-driven learning methods based on deep neural network models have been recently shown to be quite promising for light fields super-resolution. Stacked input images are up-scaled to a target resolution using  bicubic interpolation and super-resolved using a spatial convolutional neural network (CNN) in \cite{Yoon_2015} which learns a non-linear mapping between low- and high-resolution views. The output of the spatial CNN is then fed into a second CNN to perform angular super-resolution. While the approach in \cite{Yoon_2015}
takes at the input of the spatial CNN pairs or $4$-tuples of neighboring views, leading to three spatial CNNs to be learned, a single CNN is proposed by the same authors in \cite{Yoon2017} to process each view independently. 
The problem of angular super-resolution of light fields is also addressed in \cite{Kalantari2016} using an architecture based on two CNNs, one CNN being used to estimate disparity maps and the second CNN being used to synthesis intermediate views. The authors in \cite{Wu_2017b} define a CNN architecture in the EPI to increase the angular resolution.

One can also cite some related work using a hybrid light field imaging system coupling a high-resolution camera with either a light field camera
\cite{wang2017light} or with multiple low-resolution cameras \cite{Wang2017}.
In \cite{wang2017light}, the HR image captured by the DSLR camera is used to super-resolve the low-resolution images captured by an Illum light field camera. 
The authors in \cite{Wang2017} describe an acquisition device formed by eight low-resolution side cameras arranged around a central high-quality SLR lens.
A super-resolution method, called iterative patch- and depth-based synthesis (iPADS), is then proposed to reconstruct a light field with the spatial resolution of the SLR camera and an increased number of views. 

In this paper, we propose a spatial light field super-resolution method using a deep CNN (DCNN) with ten convolutional layers. Instead of using DCNN to restore each sub-aperture image independently, as done in \cite{Yoon_2015,Yoon2017}, we restore all sub-aperture images within a light field simultaneously.
This allows us to exploit both spatial and angular information to restore the light field and thus generate light fields which are angularly coherent.
A Na\"ive approach is to train a DCNN with $n = P \times Q$ inputs, where $P$ and $Q$ represent the number of vertical and horizontal angular views respectively.
However, this will significantly increase the complexity of the DCNN which makes it harder to train and more prone to over-fitting .
Instead, given that each sub-aperture image captures the same scene from a different view point, we align all sub-aperture images to the centre view using optical flow and then reduce the angular dimension of the aligned light field using a low-rank model of rank $k$, where $k \ll n$. Results in section \ref{sec:dimensionality_reduction} show that the alignment allows us, with the considered low rank model, to significantly reduce the angular dimension  of the light field.
The linearly independent column-vectors of the low-rank representation of the aligned light field, which constitute an embedding of the light field views in a lower-dimensional space, are then considered as a volume and simultaneously restored using a DCNN with $k$ input channels.
This allows us to significantly reduce the complexity of the network which is easier to train while still preserving angular consistency.
The restored column-vectors are then  combined to reconstruct the aligned high-resolution light field.
In the final stage we use inverse warping to restore the original disparities of the light field and fill the cracks caused by occlusion using a novel diffusion based inpainting strategy that propagates the restored pixels along the dominant orientation of the EPI.

Simulation results demonstrate that the proposed method outperforms all other schemes considered here when tested on 13 different light fields from two different datasets. It is important to mention that our method was not trained on the Stanford light fields and these results clearly show that our proposed method generalizes well even when considering light field structures whose disparities are significantly larger than those used for training.
Further analysis in section \ref{sec:results} shows that additional gain in performance can be achieved using iterative back projection (IBP) as a post processing step.
These results show that our method can significantly outperform existing light field super-resolution methods including the deep learning-based light field super-resolution method presented in \cite{Yoon2017}.

This paper is organized as follows. After introducing the notation in section \ref{sec:notation}, we describe the proposed method in section \ref{sec:proposed_method}. Section \ref{sec:results} discusses the simulation results with different types of light fields and provide the final concluding remarks in section \ref{sec:conclusion}.

\section{Notation and Problem Formulation}
\label{sec:notation}

We consider here the simplified 4D representation of light fields called 4D light field in \cite{Levoy_1996} and lumigraph in \cite{Gortler1996}, describing
the radiance along rays by a function $I(x,y,s,t)$ where the pairs $(x,y)$ and $(s,t)$ respectively represent spatial and angular coordinates.
The light field can be seen as capturing an array of viewpoints (called sub-aperture images) of the scene with varying angular coordinates $(s,t)$. 
The different views will be denoted here by $\mathbf{I}_{s,t} \in \mathbb{R}^{X,Y}$, where $X$ and $Y$ represent the vertical and horizontal dimension of each sub-aperture image. 

In the following, the notation $\mathbf{I}_{s,t}$ for the different sub-aperture images will be simplified as $\mathbf{I}_i$ with a bijection between $(s,t)$ and $i$. The complete light field can hence be represented by a matrix $\mathbf{I} \in \mathbb{R}^{m,n}$:
\begin{equation}
\label{eq:light_field_notation}
\mathbf{I} = \left[ vec(\mathbf{I}_1) \ | \ vec(\mathbf{I}_2) \ | \ \cdots \ | \ vec(\mathbf{I}_n) \right]
\end{equation}

\noindent with $vec(\mathbf{I}_i)$ being the vectorized representation of the $i-$th sub-aperture image, $m$ represents the number of pixels in each view ($m = X \times Y$) and $n$ is the number of views in the light field ($n = P \times Q$), where $P$ and $Q$ represent the number of vertical and horizontal angular views respectively.

Let $\mathbf{I}^H$  and $\mathbf{I}^L$ denote the high- and low- resolution light fields, respectively. 
The super-resolution problem can be formulated in Banach space as

\begin{equation}
	\label{eq:inverse_problem}
	\mathbf{I}^L = \mathbf{\downarrow}_\alpha \mathbf{B} \mathbf{I}^H+ \boldsymbol{\eta}
\end{equation}

\noindent where  $\boldsymbol{\eta}$ is an additive noise matrix, $\mathbf{\downarrow}_\alpha$ is a downsampling operator applied on each sub-aperture image where $\alpha$ is the magnification factor and $\mathbf{B}$ is the blurring kernel. There are many possible high-resolution light fields $\mathbf{I}^H$ which can produce the input low-resolution light field $\mathbf{I}^L$ via the acquisition model defined in \eqref{eq:inverse_problem}. Hence, solving this ill-posed inverse problem requires introducing some priors on $\mathbf{I}^H$, which can be a statistical prior such as a GMM model \cite{Mitra2012}, or priors learned from training data as in \cite{Yoon_2015,Yoon2017, Farrugia2017}. 

Another way to visualize a light field is to consider the EPI representation.
An EPI is a spatio-angular slice from the light field, obtained by fixing one of the spatial coordinates and one of the angular coordinates.
Consider we fix $y := y\ast$ and $t := t\ast$, an EPI is an image defined as $\boldsymbol{\epsilon}_{y\ast,t\ast} := \mathbf{I}(x,y\ast,s,t\ast)$.
Alternatively, the vertical EPI is obtained by fixing $x := x\ast$ and $s := s\ast$.
Figure \ref{fig:subjective_Eval_diff_inpaint} shows a typical EPI structure, where the slopes of the isophote lines in the EPI are related to the disparity between the views \cite{Wanner2014}.
Isophote lines with a slope of \rad{\pi/2} indicate that there is no disparity across the views while the larger is the difference between the slope and \rad{\pi/2} the larger is the disparity across the views.

\section{Proposed method}
\label{sec:proposed_method}

\begin{figure*}
\centering
\includegraphics[width=18cm]{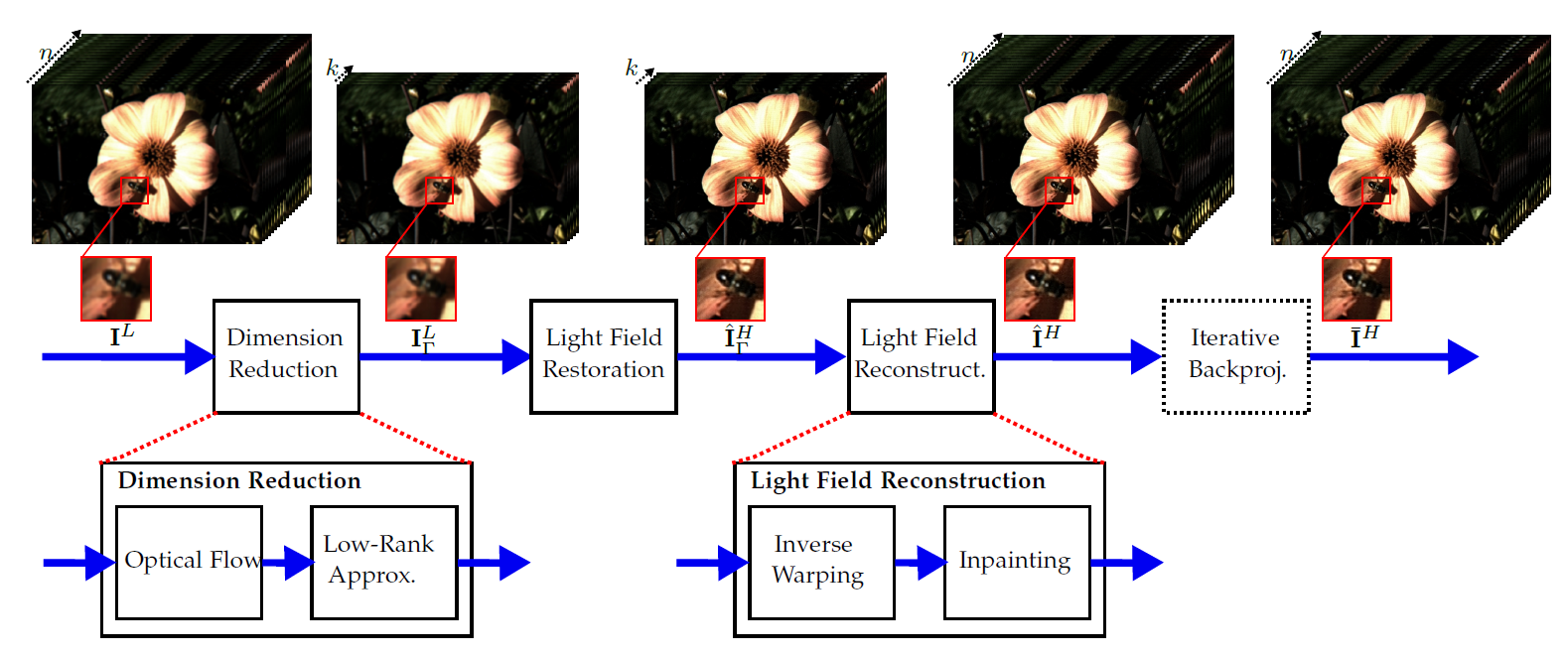}
\caption{Block diagram of the proposed light field super-resolution method}
\label{fig:block_diagram}
\end{figure*}

Figure \ref{fig:block_diagram} depicts the block diagram of the proposed spatial light field super-resolution algorithm.
Each sub-aperture image of the low-resolution light field $\mathbf{I}^L$ is first bicubic interpolated so that both $\mathbf{I}^H$ and $\mathbf{I}^L$ have the same resolution.
While a light field consists of a very large volume of high-dimensional data, it also contains a lot of redundant information since every sub-aperture image captures the same scene from a different viewpoint.
Moreover, different light field capturing devices have different spatial and angular specifications, which makes it very hard for a learning-based algorithm to learn a generalized model suitable to restore all kind of light fields irrespective of the capturing device. The \textit{Dimensionality Reduction} module tries to solve both problems simultaneously where it uses optical flow to align the light field and a low-rank matrix approximation to reduce the dimension of the light field. 
Results in section \ref{sec:dimensionality_reduction} show that we can reduce the dimensionality of the light field from $\mathbb{R}^{m,n}$ to $\mathbb{R}^{m,k}$, where $k \ll n$ is the rank of the matrix, while preserving most of the information contained in the light field.

The \textit{Light Field Restoration} module then considers the $k$ linear independent column-vectors of the rank-$k$ representation of the low-resolution light field as an embedding of the light field. We then use a DCNN to recover the texture details of the light field embedding in the lower dimensional space. The super-resolved embedding gives an estimate of the aligned high-resolution light field.
The \textit{Light Field Reconstruction} module then warps the estimated aligned high-resolution light field to restore the original disparities. 
Holes corresponding to cracks or occlusions are then filled in by diffusing information in the Epipolar Plane Images (EPI) along directions of isophote lines computed, for the positions of missing pixels, in the EPI of the low-resolution light field.
\textit{Iterative back-projection} can be further used as a post-process to refine the super-resolved light field and assure to be consistent with the low-resolution light field. 
More information about each module is provided in the following subsections.

\subsection{Light Field Dimensionality Reduction}
\label{sec:dimensionality_reduction}

Many acquisition devices have been recently designed to capture light fields, including multi-sensor approaches   \cite{Wilburn2005}, time-sequential capture methods \cite{Kim2013,Taguchi2010} and plenoptic cameras  \cite{Ng2005, Veeraraghavan2007}. 
All these different light field cameras have diverse spatial and angular specifications, which makes it very hard for a learning-based algorithm to learn a generalized model suitable to restore any kind of light field independent from the source.
Moreover, a light field contains a huge amount of redundant information since it represents a different view of the same scene.

\begin{figure*}[!h]
\centering
\begin{tabular}{cccccccc}
\centering

\footnotesize{{No Align}} & \footnotesize{{RASL}} \cite{peng_2010} &  \footnotesize{LRMA \cite{Jiang2017}} & \footnotesize{BMA} &\footnotesize{Horn-Scunk \cite{horn_1980}}& \footnotesize{SIFT Flow \cite{liu_2011}} & \footnotesize{CPM \cite{hu_2016}} & \footnotesize{SPM-BP \cite{li_2015}} \\ 
\includegraphics[width=1.9cm]{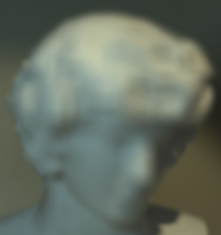} & 
\includegraphics[width=1.9cm]{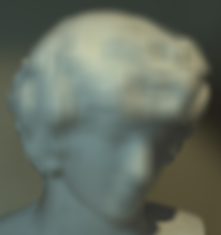} & 
\includegraphics[width=1.9cm]{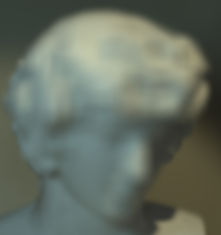}
&
\includegraphics[width=1.9cm]{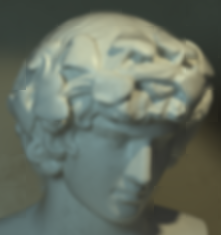}
&
\includegraphics[width=1.9cm]{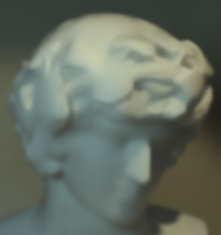}  
&
\includegraphics[width=1.9cm]{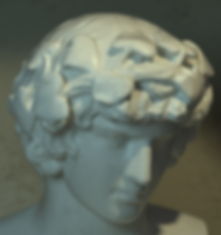} 
&
\includegraphics[width=1.9cm]{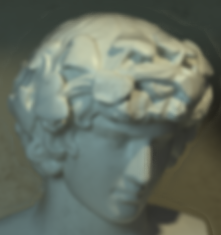} 
&
\includegraphics[width=1.9cm]{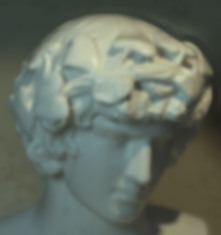} 
\\
\footnotesize{44.714} & \footnotesize{43.896} & \footnotesize{44.787} &\footnotesize{4.434} & \footnotesize{16.466} & \footnotesize{1.796} & \footnotesize{7.809} & \footnotesize{5.817} 
\\
\includegraphics[width=1.9cm]{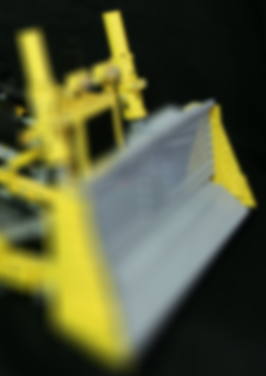} & 
\includegraphics[width=1.9cm]{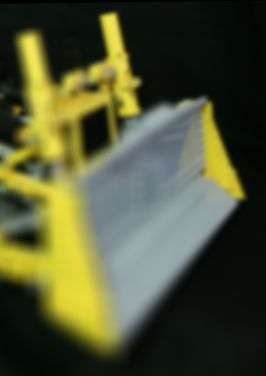} & 
\includegraphics[width=1.9cm]{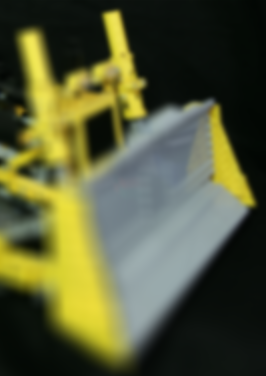}
&
\includegraphics[width=1.9cm]{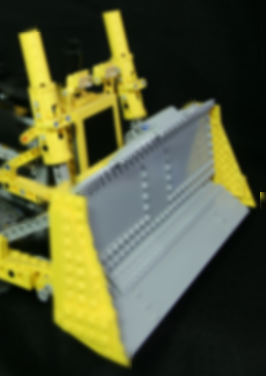}
&
\includegraphics[width=1.9cm]{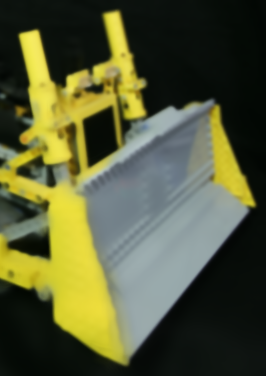}  
&
\includegraphics[width=1.9cm]{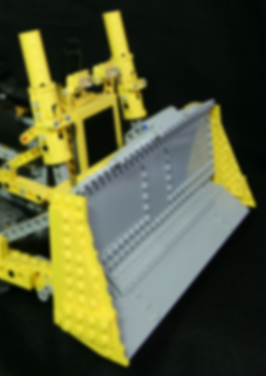} 
&
\includegraphics[width=1.9cm]{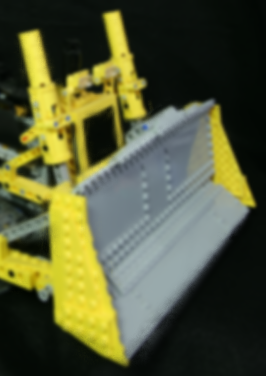} 
&
\includegraphics[width=1.9cm]{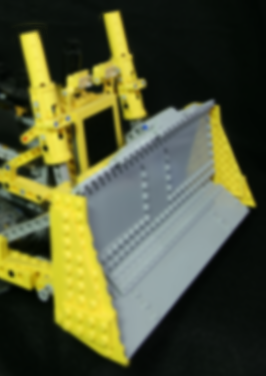} 
\\
\footnotesize{306.960} & \footnotesize{286.078} & \footnotesize{306.166} & \footnotesize{24.519} & \footnotesize{45.588} & \footnotesize{8.560} & \footnotesize{68.067} & \footnotesize{22.313}
\\ 
\includegraphics[width=1.9cm]{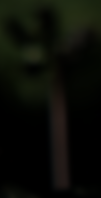} & 
\includegraphics[width=1.9cm]{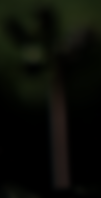} 
& 
\includegraphics[width=1.9cm]{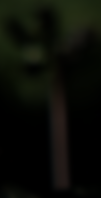} 
&
\includegraphics[width=1.9cm]{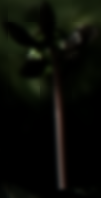} 
&
\includegraphics[width=1.9cm]{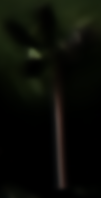}  
&
\includegraphics[width=1.9cm]{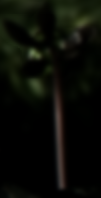} 
&
\includegraphics[width=1.9cm]{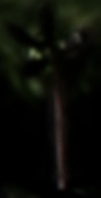} 
&
\includegraphics[width=1.9cm]{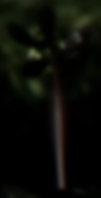}  \\
\footnotesize{99.092} & \footnotesize{98.261} & \footnotesize{99.080} & \footnotesize{46.374} & \footnotesize{36.388} & \footnotesize{19.164} & \footnotesize{51.073} & \footnotesize{71.652} \\ 
\end{tabular} 
\caption{Cropped regions of the mean sub-aperture images when using different disparity compensation methods. Underneath each image we provide the average variance across the $n$ sub-aperture images which was used in \cite{Farrugia2017} to characterize the performance of the alignment algorithm, where smaller values indicate  better alignment.  }
\label{fig:subjective_Eval1}
\end{figure*}

The authors in \cite{Jiang2017} hypothesized that this redundancy can be suppressed by jointly aligning the sub-aperture images in the light field and estimating a low-rank approximation (LRA) of the light field.
This approach has shown very promising results in the field of light field compression.
In the same spirit, the RASL algorithm \cite{peng_2010} was used to find the homographies that globally align a batch of linearly correlated images.
Both methods seek for an optimal set of homographies such that the matrix of aligned images can be decomposed in a low-rank matrix of aligned images, with the latter constraining the error matrix to be sparse.
The results in Figure \ref{fig:subjective_Eval1} show that while both RASL and LRA methods manage to globally align the sub-aperture images, the mean sub-aperture image is still very blurred, indicating that the light field is not suitably aligned.

The authors in \cite{Farrugia2017} have used the block matching algorithm (BMA) to align patch volumes. The results in Figure \ref{fig:subjective_Eval1} show that BMA manages to align better the sub-aperture images, where the average variance across the $n$ sub-aperture image is significantly reduced.
This result suggests that local methods can improve the alignment of the sub-aperture images, which as we will see in the sequel, will allow us to significantly reduce the dimensionality of the light field.

In this paper, we formulate the light field dimensionality reduction problem as
\begin{equation}
\label{eq:dim_reduction_formulation}
\underset{\mathbf{u},\mathbf{v},\mathbf{A}}{
min}_{}{ || \Gamma_{\mathbf{u},\mathbf{v}} \left( \mathbf{I}^L\right) - \mathbf{A}||^2_2 \quad \text{ s.t. } \quad rank(\mathbf{A}) = k}
\end{equation}

\noindent where $\mathbf{u} \in \mathbb{R}^{m,n}$ and $\mathbf{v} \in \mathbb{R}^{m,n}$ are flow vectors that specify the displacement of each pixel needed to align each sub-aperture image with the centre view, $\mathbf{A}$ is a  rank-$k$ matrix which approximates the aligned light field and $\Gamma_{\mathbf{u},\mathbf{v}} \left( \cdot \right)$ is a forward warping operator (which here performs a disparity compensation where the disparities maps $(\mathbf{u},\mathbf{v})$ are estimated with an optical flow estimator).
This optimization problem is computationally intractable.
Instead, we decompose this problem in two sub-problems: \romannumeral 1) use optical flow to find the flow matrices $\mathbf{u} $ and $\mathbf{v}$ that best align each sub-aperture image with the centre view and \romannumeral 2) use low-rank approximation to derive the rank-$k$ matrix that minimizes the error with respect to the aligned light field.

\subsubsection{Optical Flow}
\label{sec:optical_flow}

The problem of aligning all the sub-aperture images with the centre view can be formulated as

\begin{equation}
\label{eq:optical_flow}
\mathbf{I}_j(x,y) = \mathbf{I}_i(x + \mathbf{u}_i,y + \mathbf{ v}_i) \quad i \in \left[1,n \right] (i \neq j)
\end{equation}

\noindent where $j$ corresponds to the index of the centre view, and $\left( \mathbf{u}_i, \mathbf{v}_i \right)$ are the flow vectors optimal to align the $i$-th sub-aperture image with the centre view.
There are several optical flow algorithms intended to solve this problem \cite{horn_1980,liu_2011,li_2015,hu_2016} where Figure \ref{fig:subjective_Eval1} shows the performance of some of these methods.
It can be seen that the mean aligned sub-aperture images computed using \cite{horn_1980} is generally blurred while those aligned using the methods in \cite{li_2015, hu_2016} generally provide ghosting artefacts at the edges.
Moreover, it can be seen that SIFT Flow \cite{liu_2011} generally provides very good alignment and manages to attain the smallest variation across the sub-aperture images. 
While the SIFT Flow algorithm will be used in this paper to compute the flow vectors, any other optical flow method can be used.

\subsubsection{Low-Rank Approximation}
\label{sec:lra}

Given that the flow-vectors $\left( \mathbf{u}_i, \mathbf{v}_i \right)$ for the $i$-th sub-aperture image are already available, the minimization problem in Eq. \eqref{eq:dim_reduction_formulation} can now be reduced to

\begin{equation}
\label{eq:simple_dim_reduction_formulation}
\underset{\mathbf{\mathbf{B}^L, \mathbf{C}^L}}{
min}_{}{ || \mathbf{I}^L_\Gamma - \mathbf{B}^L\mathbf{C}^L||^2_2} \quad \text{ s.t. } \quad rank(\mathbf{B}^L) = k
\end{equation}

\noindent where $\mathbf{I}^L_\Gamma = \Gamma_{\mathbf{u},\mathbf{v}} \left( \mathbf{I}^L\right)$, $\mathbf{B}^L \in  \mathbb{R}^{m,k}$ is a  rank-$k$ matrix and $\mathbf{C}^L \in  \mathbb{R}^{k,n}$ is the combination weight matrix.
These matrices can be found using  singular value decomposition (SVD) $\mathbf{I}^L_\Gamma = \mathbf{U} \boldsymbol{\Sigma} \mathbf{V}^T$, where 
$\mathbf{B}^L$ is set as the $k$ first columns of $\mathbf{U} \boldsymbol{\Sigma}$ and $\mathbf{C}^L$ is set as the $k$ first rows of $\mathbf{V}^T$, so that $\mathbf{B}^L\mathbf{C}^L$ is the closest $k$-rank approximation of the aligned light field $\mathbf{I}^L_\Gamma$. The error matrix $\mathbf{E}^L$ is the error matrix which is simply computed using $\mathbf{E}^L = \mathbf{I}^L_\Gamma - \mathbf{B}^L\mathbf{C}^L$

\begin{figure*}
\begin{subfigure}{.5\textwidth}
	\includegraphics[width=.95\linewidth]{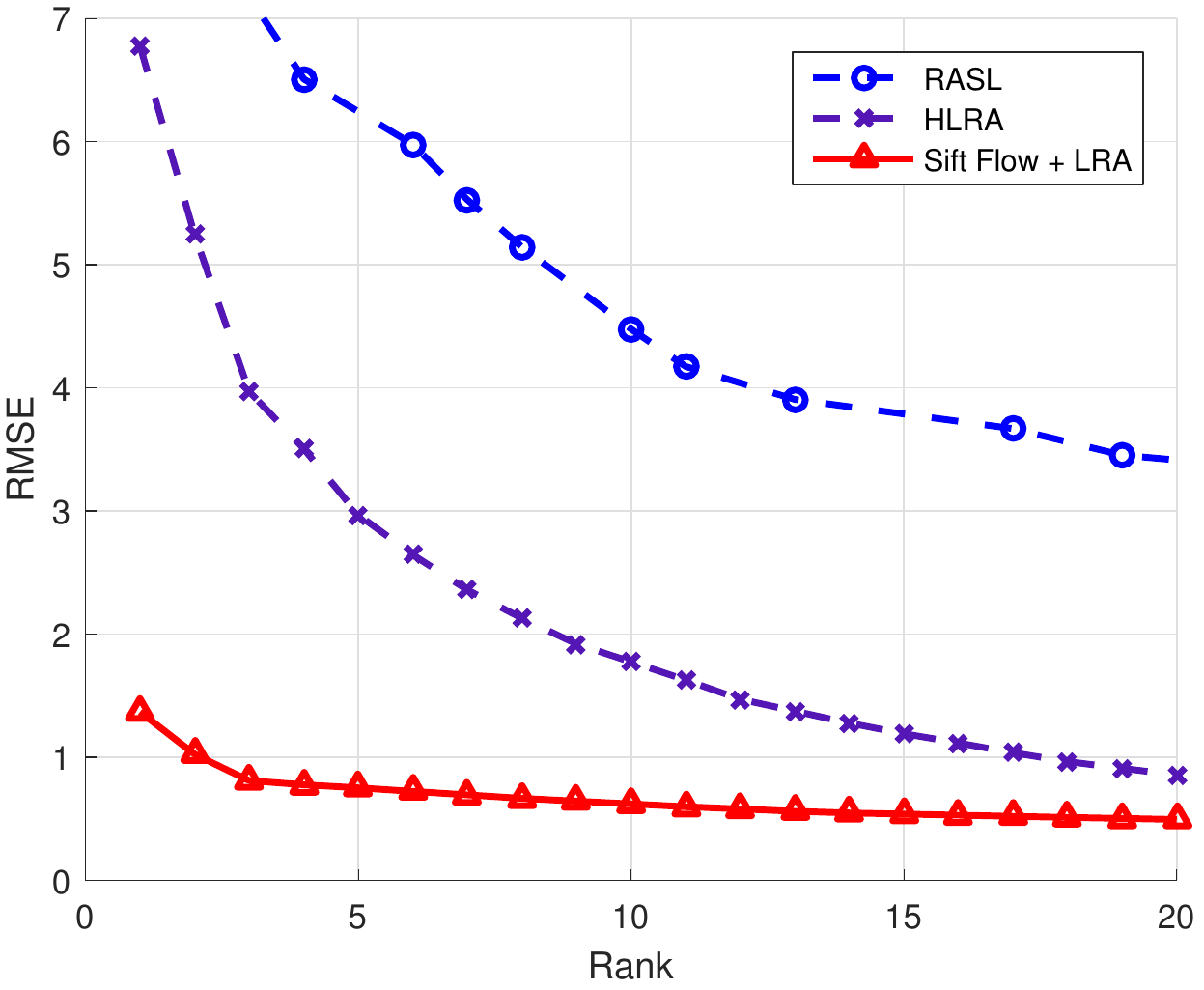}
\end{subfigure}
\begin{subfigure}{.5\textwidth}
	\includegraphics[width=.95\linewidth]{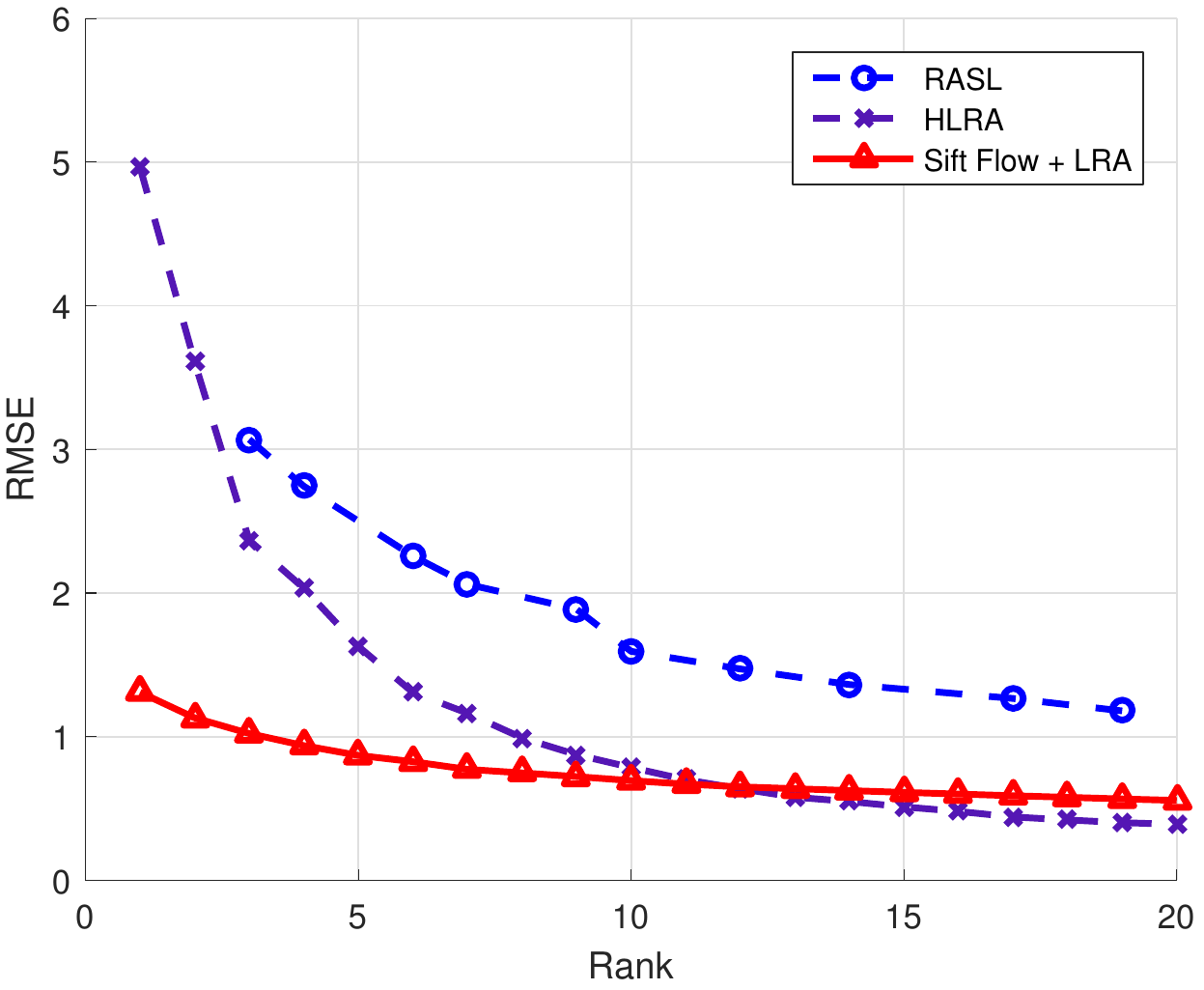}
\end{subfigure}

\begin{tabular}{@{}c@{}c@{}c c@{}c@{}c}
\centering
\includegraphics[width=2.9cm]{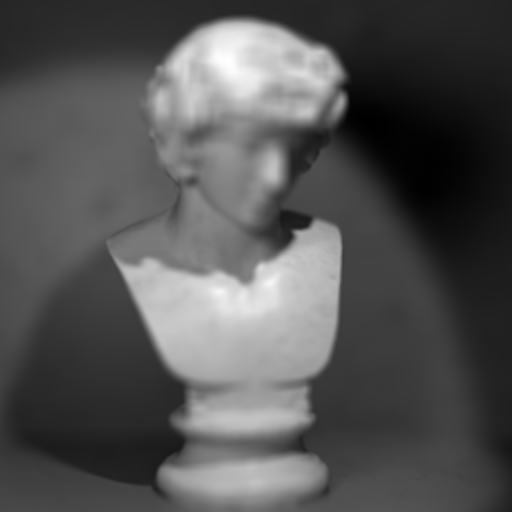} & 
\includegraphics[width=2.9cm]{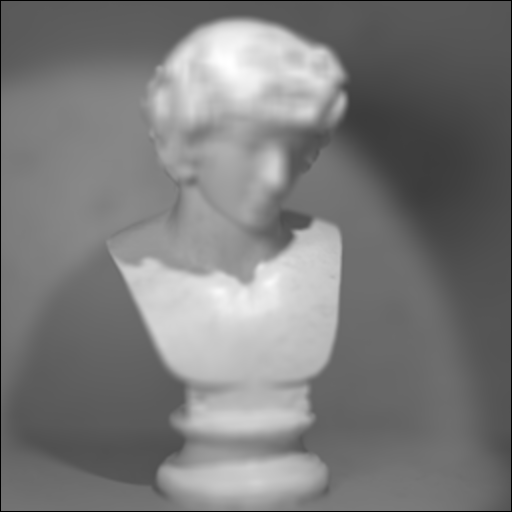} & \includegraphics[width=2.9cm]{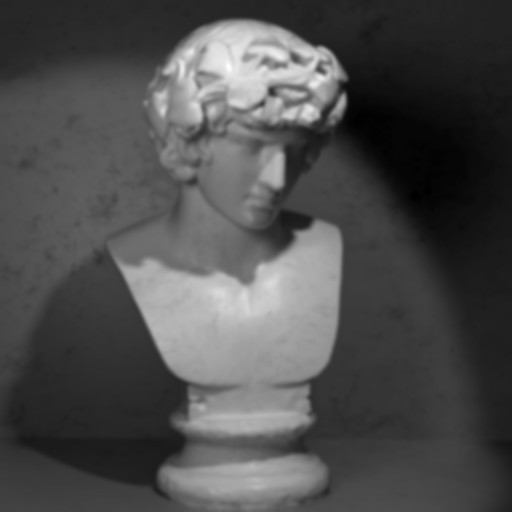} & 
\includegraphics[width=2.9cm]{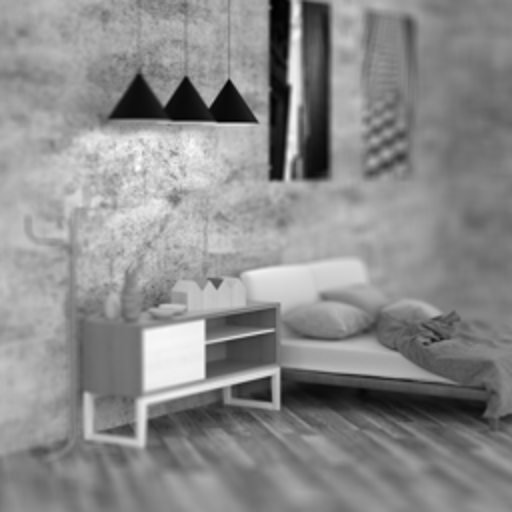} &
\includegraphics[width=2.9cm]{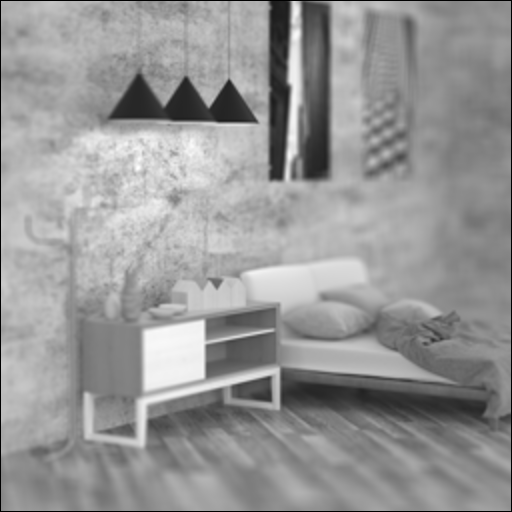} &
\includegraphics[width=2.9cm]{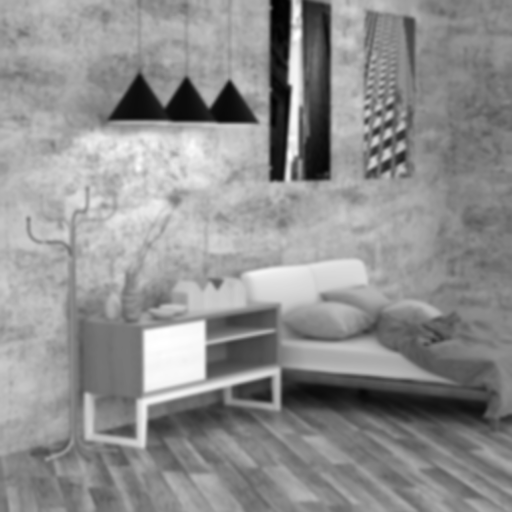}\\
\footnotesize{PCA} & \footnotesize{HLRA} & \footnotesize{Sift Flow + LRA} & \footnotesize{PCA} & \footnotesize{HLRA} & \footnotesize{Sift Flow + LRA}\\
\end{tabular}

\caption{These figures show how the error between the low-rank and full rank representation vary at different ranks. It can be seen that using optical flow to align the light field followed by low-rank approximation attains the best performance. The images in the second row show the principal basis derived using different methods. The sharper the principal basis is the more information is being captured in the principal basis.}
\label{fig:low_rank_analysis}
\end{figure*}

Figure \ref{fig:low_rank_analysis} depicts the performance of three different dimensionality reduction techniques at different ranks. 
To measure the dimensionality reduction ability of these methods we compute the root mean square error (RMSE) between the aligned original and the rank-$k$ representation of the aligned light field.
It can be seen that the RASL algorithm has the largest distortions at almost all ranks when compared to the other two approaches. On the other hand, it can be seen that HLRA manages to significantly outperform the RASL method, which gain is attained by the improved alignment. 
Nevertheless, it can be clearly observed that the proposed Sift Flow + LRA method gives the best performance, especially at lower ranks, indicating that more information is captured within the low-rank matrix.
To emphasize this point we show in figure  \ref{fig:low_rank_analysis} the principal basis of PCA, HLRA and Sift Flow + LRA. PCA is computed on the light field without disparity estimation and therefore can be considered here as a baseline to show that alignment allows us to get more information in the principal basis.
Moreover, it can be seen that the principal basis derived using our Sift Flow + LRA manages to capture more texture detail in the principal basis than the other methods and confirms the benefit that local alignment has on the energy compaction ability of the proposed dimensionality reduction method.\footnote{Note that the RASL method does decompose the matrix into a combination of basis elements and therefore the principal basis of RASL could not be shown here. }

\subsection{Light Field Restoration}
\label{sec:restoration}

We consider a low-rank representation of the aligned low-resolution light field $\mathbf{A}^L = \mathbf{B}^L \mathbf{C}^L$, where $\mathbf{A}^L \in \mathbb{R}^{m,n}$  is a rank-$k$ matrix with $k \ll n$.
Similarly, $\mathbf{A}^H = \mathbf{B}^H \mathbf{C}^H$ is a rank-$k$ representation of the aligned high-resolution light field.
The rank of a matrix is defined as the maximum number of linearly independent column vectors in the matrix. 
Moreover, the linearly dependent column vectors of a matrix can be reconstructed using a weighted summation of the linearly independent column vectors of the same matrix.
This leads us to decompose $\mathbf{A}^L$ in two sub-matrices: $\breve{\mathbf{A}}^L \in \mathbb{R}^{m,k}$ which groups the linear independent column vectors of $\mathbf{A}^L$ and $\hat{\mathbf{A}}^L \in \mathbb{R}^{m,n-k}$ which groups the linearly dependent column vectors of $\mathbf{A}^L$. In practice, we decompose the rank-$k$ matrix $\mathbf{A}^L$ using QR decomposition (\textit{i.e}. $\mathbf{A}^L = \mathbf{Q} \mathbf{R}$). The index of the linearly independent components of $\mathbf{A}^L$ then correspond to the index of the non-zero diagonal elements of the upper-triangular matrix $\mathbf{R}$.
We then use the same indices to decompose $\mathbf{A}^H$ into sub-matrices $\breve{\mathbf{A}}^H$ and $\hat{\mathbf{A}}^H$. The matrix $\hat{\mathbf{A}}^L$ can be reconstructed as a linear combination of $\breve{\mathbf{A}}^L$, where the weight matrix $\mathbf{W}$ is computed using
\begin{equation}
\mathbf{W} = \left ( \breve{\mathbf{A}}^{L\intercal}\breve{\mathbf{A}}^L \right )^\dagger \breve{\mathbf{A}}^{L\intercal}\hat{\mathbf{A}}^L
\end{equation}

\noindent where $\left ( \cdot \right)^\dagger$ stands for the pseudo inverse operator. We assume here that the weight matrix $\mathbf{W}$,  which is optimal in terms of least squares to reconstruct 
$\hat{\mathbf{A}}^L$, is suitable to reconstruct $\hat{\mathbf{A}}^H$.

\begin{figure*}
\centering
\includegraphics[width=18cm]{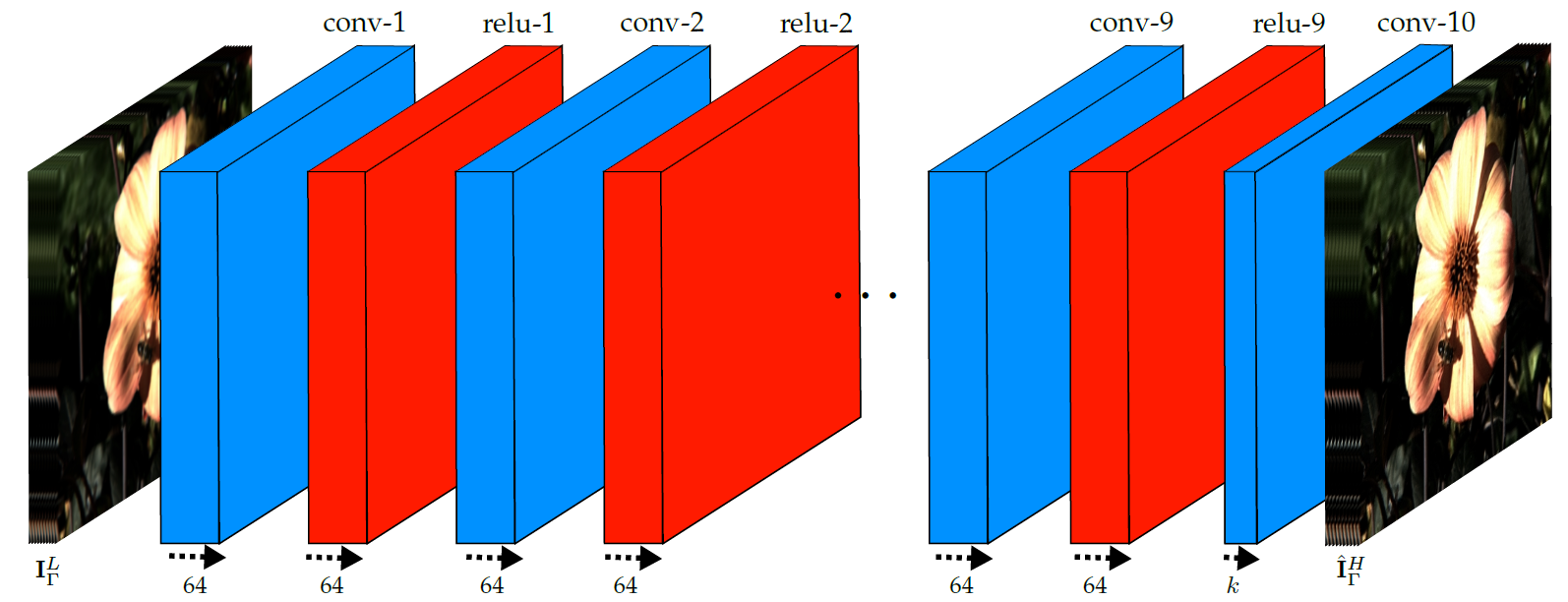}\\
\caption{The proposed network structure which receives a low-resolution light field and restores it using the proposed DCNN.}
\label{fig:cnn_architecture}
\end{figure*}

Driven by the recent success of deep learning in the field of single-image \cite{dong_2013,kim_2016} and light field super-resolution \cite{Yoon_2015,Yoon2017}, we use a DCNN to model the upscaling function that minimizes the following objective function

\begin{equation}
\sfrac{1}{2} ||\breve{\mathbf{A}}^H - f\left( \breve{\mathbf{A}}^L \right) ||^2
\end{equation}

\noindent where  $f(\cdot)$ is a function modelled by the DCNN illustrated in Figure \ref{fig:cnn_architecture} which has ten convolutional layers.
The linearly independent sub-matrix $\breve{\mathbf{A}}^L$ is passed through a stack of convolutional and rectified linear unit (ReLU) layers.
We use a convolution stride of 1 pixel with no padding nor spatial pooling.
The first convolutional layer has 64 filters of size $3 \times 3 \times k$ while the last layer, which is used to reconstruct the high-resolution light field, employs $k$ filter of size $3 \times 3 \times 64$. All the other layers use 64 filters of size $3 \times 3 \times 64$ which are initialized using the method in \cite{xavier_2010}. The DCNN was trained  using a total of 200,000 random patch-volumes of size $64 \times 64 \times k$ from the 98 low- and high-resolution low-rank approximation of rank $k$ of the light fields from the EPFL, INRIA and HCI datasets\footnote{It must be noted that none of the light fields used for validation were used for training.}.
The Titan GTX1080Ti Graphical Processing Unit (GPU) was used to speed up the training process.

During the evaluation phase, we estimate the super-resolved linearly independent representation of the light field $\breve{\mathbf{A}}^H = f \left( \breve{\mathbf{A}}^L\right)$. We then estimate the super-resolved linear dependent part of the light field using
\begin{equation}
\hat{\mathbf{A}}^H = \breve{\mathbf{A}}^H \mathbf{W}
\end{equation}

\noindent The super-resolved low-rank representation of the aligned light field is then derived by the union of the two matrices $\hat{\mathbf{A}}^H$ and $\breve{\mathbf{A}}^H$ \textit{i.e}. $\mathbf{A}^H = \hat{\mathbf{A}}^H \cup \breve{\mathbf{A}}^H$. The super-resolved light field is then reconstructed using
\begin{equation}
\hat{\mathbf{I}}^H_\Gamma = \mathbf{A}^H + \mathbf{E}^L
\end{equation}

\subsection{Light Field Reconstruction}
\label{sec:light_field_reconstruction}

\begin{figure}[!h]
\centering
\begin{subfigure}{.5\textwidth}

\begin{tabular}{ccc}
\centering

\includegraphics[width=2.6cm]{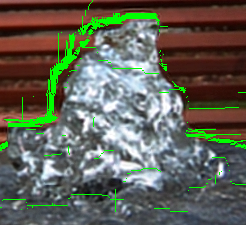} &
\includegraphics[width=2.6cm]{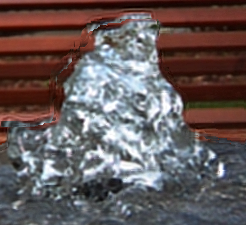} 
&
\includegraphics[width=2.6cm]{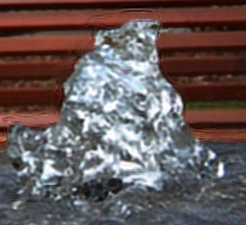} \\
\includegraphics[width=2.6cm]{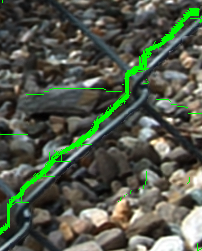} &
\includegraphics[width=2.6cm]{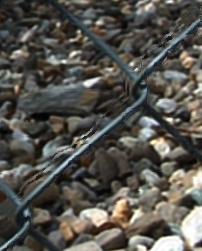} 
&
\includegraphics[width=2.6cm]{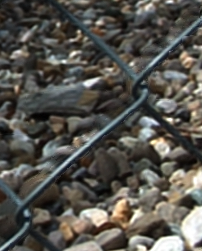} \
\\
\includegraphics[width=2.6cm]{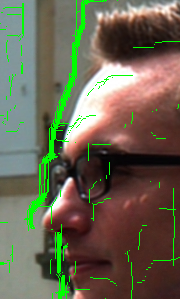} &
\includegraphics[width=2.6cm]{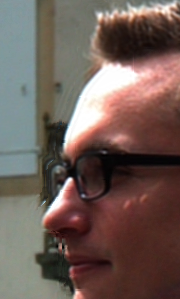} 
&
\includegraphics[width=2.6cm]{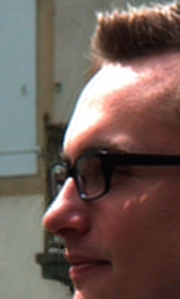} \\
\footnotesize{Forward Warping} & \footnotesize{Bicubic Interpolation} & \footnotesize{EPI Diffusion}\\ 
\end{tabular} 
\caption{Inpainting the cracks marked in green}
\label{fig:subjective_Eval2}
\end{subfigure}
\begin{subfigure}{.5\textwidth}
\includegraphics[width=8.8cm]{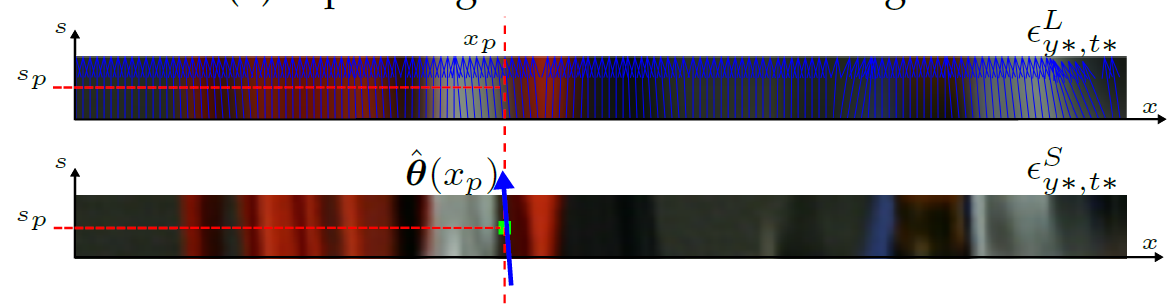} \\
\caption{Diffusion based inpainting}
\label{fig:subjective_Eval_diff_inpaint}
\end{subfigure}
\caption{Filling the missing information caused by occlusion.}
\end{figure}

The restored aligned light field $\hat{\mathbf{I}}^H_\Gamma$ has all sub-aperture images aligned with the centre view. 
A na\"ive approach to recover the original disparities of the restored sub-aperture images is to use forward warping.
However, as can be seen in the first column of Figure \ref{fig:subjective_Eval2}, forward warping is not able to restore all pixels and results in a number of cracks or holes corresponding to occlusions (marked in green).
One can use either bicubic interpolation or inverse warping to fill the holes.
However, in case of occlusions, the neighbouring pixels may not be well correlated with the missing information, which often results in inaccurate estimations (see Figure \ref{fig:subjective_Eval2} second column). 
More advanced inpainting algorithms \cite{criminisi_2004,xu_2010} can be used to restore each hole separately.
However, these methods do not exploit the light field structure and therefore provide inconsistent reconstruction of the same spatial region at different angular views.

In this work, we use a diffusion based inpainting algorithm that estimates the missing pixels by diffusing information available in other views.
Similar to the work in \cite{frigo_2017}, we exploit the EPI structure to diffuse information along the dominant orientation of the EPI. 
However, instead of predicting the orientation of unknown pixels from their spatial neighborhood as done in \cite{frigo_2017}, we exploit the similarity between the low- and super-resolved EPIs and use the structure tensor computed on the low-resolution EPI to guide the inpainting process in the high-resolution EPI.

Without loss of generality we consider the EPI where the dimensions $y \ast$ and $t \ast$ are fixed.
The case of vertical slices is analogous. 
We first compute the structure tensor of the low-resolution EPI $\boldsymbol{\epsilon}^L_{y \ast, t \ast}$ at coordinates $(x,s)$ using

\begin{equation}
\mathbf{T}\left ( x,s\right ) = \nabla \boldsymbol{\epsilon}^L_{y \ast, t \ast}\left ( x,s\right ) \nabla \boldsymbol{\epsilon}^L_{y \ast, t \ast}\left ( x,s\right )^\intercal
\label{eq:structure_tensor}
\end{equation}

\noindent where $\nabla$ stands for the single order gradient computed using the sobel kernel. 
The authors in \cite{frigo_2017} compute an average weighting of the columns of $\mathbf{T}\left ( x,s\right )$ to derive the dominant orientation, where the weights are given by an anisotropy measure.
Nevertheless, the anisotropy may fail in regions that are smooth and therefore the weighted average may fail in these regions to estimate the dominant direction of the EPI.
This problem becomes an issue when computing these orientations on low-resolution versions of the light fields.
Instead, we estimate the orientation at every pixel in the EPI by computing the eigen decomposition of $\mathbf{T}\left ( x,s\right )$ and choose the direction $\boldsymbol{\theta}\left( x, s\right)$ which corresponds to the eigen-vector with the smallest eigen-value.
Moreover, driven by the observation that the disparities in a light field are typically small, and considering that the local slope in the EPI is proportional to the disparity, it is reasonable to assume that slopes which correspond to large disparities are less probable to occur. 
Therefore, to ensure that the tensor driven diffusion is performed along a single coherent direction per column of the EPI and reduce noise, the dominant orientation $\boldsymbol{\theta} \left ( x \right )$ is computed using the column-wise median of $\boldsymbol{\theta}\left ( x,s\right )$ whose orientation is in the range $[\frac{\pi}{2} - \alpha,  \frac{\pi}{2} + \alpha]$ radians. 
In all our experiments we set $\alpha = $ \rad{\pi/4}. 
While the dominant orientation vectors are less noisy, we further reduce the noise by applying the Total Variation (TV-L1) denoising \cite{rudin_1992} on the orientation field $\boldsymbol{\theta} \left ( x \right )$, which searches to optimize

\begin{equation}
\hat{\boldsymbol{\theta}} = arg\,min_{\hat{\boldsymbol{\theta}}}{||\nabla\hat{\boldsymbol{\theta}}||_1 + \lambda || \hat{\boldsymbol{\theta}} - \boldsymbol{\theta}||_2}
\end{equation}

\noindent where $\lambda$ is the total variation regularization parameter and was set to 0.5 in our experiments. Figure \ref{fig:subjective_Eval_diff_inpaint} (top) shows the EPI of the low-resolution light field  $\boldsymbol{\epsilon}^L_{y \ast, t \ast}$ and the dominant orientations $\hat{\boldsymbol{\theta}} \left ( x \right )$ marked by blue arrows.

The restored EPI  $\boldsymbol{\epsilon}^H_{y \ast, t \ast} := \hat{\mathbf{I}}^H(x,y\ast,s,t\ast)$ (see Figure \ref{fig:subjective_Eval_diff_inpaint} (bottom)) has a number of missing pixels (marked in green).
Consider that the hole we want to inpaint has coordinates $\left( x_p,s_p\right)$.
The aim of the proposed diffusion based inpainting algorithm is to propagate known pixels in the orientation $\hat{\boldsymbol{\theta}}\left( x_p \right)$ to fill the missing pixels.
The diffusion over the EPI $\boldsymbol{\epsilon}^H_{y \ast, t \ast}$ evolves as

\begin{equation}
\frac{\partial \boldsymbol{\epsilon}^H_{y \ast, t \ast} }{\partial s}
   = \mathrm{Tr}\hspace{1pt}\left( \hat{\boldsymbol{\theta}}\left( x_p \right)\hat{\boldsymbol{\theta}}\left( x_p \right)^\intercal \mathbf{H} \left ( x_p,s_p\right) \right)
   \label{eq:diffusion_equation}
\end{equation}

\noindent where $ \mathrm{Tr} \left( \cdot \right)$ stands for the trace operator and $\mathbf{H} \left ( x,s\right)$  denotes the Hessian of  $\boldsymbol{\epsilon}^H_{y \ast, t \ast}$ at coordinates $\left( x,s\right)$.
The term $\hat{\boldsymbol{\theta}}\left( x_p \right)\hat{\boldsymbol{\theta}}\left( x_p \right)^\intercal$ is used to enforce the diffusion to occur only in the direction of the isophote eigenvector.
The missing pixels are restored iteratively by finding the solution to \eqref{eq:diffusion_equation} which is closest to zero.
Figure \ref{fig:subjective_Eval2} (third column) shows the results attained using the proposed inpainting strategy.

In order to restore all the cracks in the light field we first fix $t \ast$ to that of the center view and iteratively restore all the horizontal EPIs for all $y \ast \in [1,Y]$ by solving Eq. \eqref{eq:diffusion_equation}. This corresponds to filling the cracks for the centre row of the matrix of sub-aperture images.
We then fix $s \ast$ to that of the centre view and iteratively restore all the vertical EPIs for all $x \ast \in [1,X]$ which effectively restores all cracks for the centre column of the matrix of sub-aperture images.
The remaining propagations are performed row-by-row where each time we restore all pixels within $t \ast \in [1,Q]$.

\subsection{Iterative Back-Projection}
\label{sec:ibp}

One problem with the method proposed in this paper is that after we restore the aligned light field $\hat{\mathbf{I}}^H_\Gamma$ we have to compute inverse warping to restore the original disparities.
However, the inverse warping is not able to recover occluded regions and some pixels are displaced by $\rpm 1$-pixel due to rounding errors.
While the former problem is solved using the method described in Section \ref{sec:light_field_reconstruction}, the second problem was not yet addressed.
Nevertheless, the results illustrated in Tables \ref{tbl:psnr_analysis_blurredx2}, \ref{tbl:psnr_analysis_blurredx3} and Figure \ref{fig:subjective_Eval_results} indicate that significant performance gains can be achieved even if we do not explicitly cater for distortions caused by rounding errors in the inverse warping process. 
Nevertheless, these distortions can be corrected using the classical method of iterative back projection \cite{irani_1991}, which is adopted by several single image super-resolution methods (see \cite{Glasner2009}) to ensure that the down sampled version of the super-resolved light field is consistent with the observed low-resolution light field.
The IBP algorithm iteratively refines the estimated high-resolution light field $\bar{\mathbf{I}}^H_\kappa$ at iteration $\kappa$ by first back-projecting it into an estimated low-resolution light field $\bar{\mathbf{I}}^L_\kappa$ using

\begin{equation}
	\label{eq:ibp1}
	\bar{\mathbf{I}}^L_\kappa =\mathbf{\uparrow}_\alpha \left ( \mathbf{\downarrow}_\alpha \mathbf{B} \bar{\mathbf{I}}^H_\kappa \right )
\end{equation}

\noindent where $\mathbf{\downarrow}_\alpha$ is a downsampling operator, $\mathbf{\uparrow}_\alpha$ is the bicubic upscaling operation, $\alpha$ is the magnification factor and $\mathbf{B}$ is the blurring kernel.
The deviation between the LR views found by back-projection
and the original LR views is then used to further correct each
HR estimated view of the light field as
\begin{equation}
	\label{eq:ibp2}
\bar{\mathbf{I}}^H_{\kappa+1} = \bar{\mathbf{I}}^H_{\kappa} + \left( \mathbf{I}^L - \bar{\mathbf{I}}^L_\kappa \right)
\end{equation}

The IBP algorithm is initiated by setting $\bar{\mathbf{I}}^H_{0} = \hat{\mathbf{I}}^H$ and the iterative procedure terminates when $\kappa = K$.
It was observed that significant improvements were achieved in the first few iterations and we therefore set $K = 10$ in our experiments. The restored light field following iterative back-projection is therefore set to $\bar{\mathbf{I}}^H = \bar{\mathbf{I}}^H_{K}$.


\section{Experimental Results}
\label{sec:results}

The experiments conducted in this paper use both synthetic and real-world light fields from publicly available datasets. We use 98 light fields from the EPFL \cite{rerabek_2016}, INRIA\footnote{INRIA dataset: \url{https://goo.gl/st8xRt}} and HCI\footnote{HCI dataset: \url{http://hci-lightfield.iwr.uni-heidelberg.de/}} for training.
We conducted the tests using light fields from the INRIA and Stanford\footnote{Stanford dataset: \url{http://lightfield.stanford.edu/}} datasets.
We use the Stanford dataset in this evaluation since it has disparities significantly larger than both INRIA and EPFL light fields, which were captured using plenoptic cameras. Moreover, unlike the HCI dataset, the Stanford light fields capture real world objects.
We therefore use this dataset to assess the generalization ability of the algorithms considered in this experiment to light fields which are captured using camera sensors which differ from the ones used for training.
While the sub-aperture images of the EPFL, HCI and Stanford datasets are available, the light fields in the INRIA dataset were decoded using the method in \cite{dansereau_2013} as mentioned on their website.
In all our experiments we consider a $9 \times 9$ array of sub-aperture images. For computational purposes, the high-resolution images of the Stanford dataset were downscaled such that the lowest dimension is set to 400 pixels.
The high-resolution images of the other datasets were kept unchanged, \textit{i.e}. $512 \times 512$ for the HCI light fields and $625 \times 434$ for both EPFL and INRIA light fields. Unless otherwise specified, the low-resolution light fields were generated by blurring each high-resolution sub-aperture image with a Gaussian filter using a window size of 7 and standard deviation of 1.6,  down-sampled to the desired resolution and up-scaled back to the target resolution using bi-cubic interpolation.
Unless otherwise specified, the iterative back-projection refinement strategy was disabled to permit a fair comparison to the other state of the art super-resolution methods considered.

We compare the performance of our system against the best performing methods found in our recent work \cite{Farrugia2017}, namely the CNN based light field super-resolution algorithm (LF-SRCNN) \cite{Yoon2017} and both linear subspace projection based methods, PCA+RR and BM+PCA+RR \cite{Farrugia2017}.
These methods were retrained using samples from the 98 training light fields mentioned above using training procedures explained in their respective papers.
Moreover, given that the very deep super-resolution (VDSR) method \cite{kim_2016} achieved state-of-the-art performance on single image super-resolution, we apply this method to restore every sub-aperture image independently.
It is important to mention here that in our previous work we found that BM+PCA+RR significantly outperforms  several other light field and single-image super-resolution algorithms including \cite{Yoon_2015,Mitra2012,peleg_2014,dong_2016,zhang_2016,
timofte_2013,dong_2013}. 
Due to space constraints we did not provide comparisons against the latter approaches.

\begin{figure*}[!h]
\centering
\begin{tabular}{ccccccc}
\centering

\footnotesize{{Original}} & \footnotesize{Bicubic}  &  \footnotesize{PCA+RR \cite{Farrugia2017}} & \footnotesize{BM+PCA+RR \cite{Farrugia2017}} & \footnotesize{LF-SRCNN \cite{Yoon2017}} &\footnotesize{VDSR \cite{kim_2016}}& \footnotesize{Proposed} \\ 
\includegraphics[width=1.9cm]{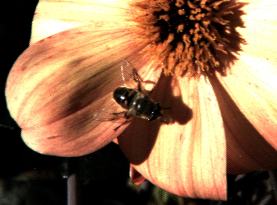} & 
\includegraphics[width=1.9cm]{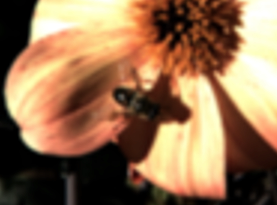} & 
\includegraphics[width=1.9cm]{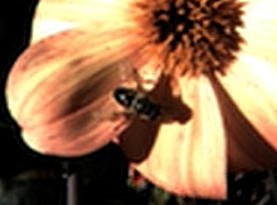}
&
\includegraphics[width=1.9cm]{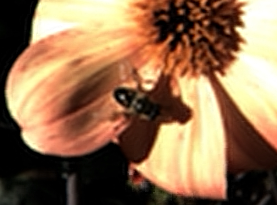}
&
\includegraphics[width=1.9cm]{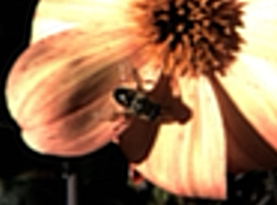}  
&
\includegraphics[width=1.9cm]{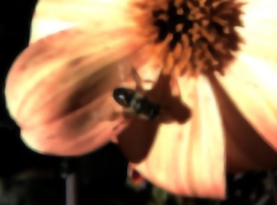} 
&
\includegraphics[width=1.9cm]{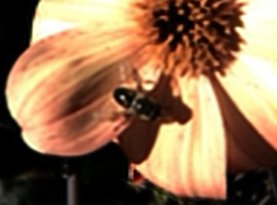} 
\\
\includegraphics[width=1.9cm]{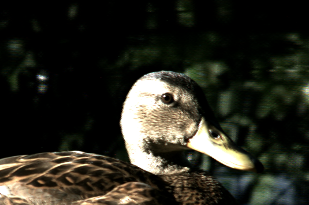} & 
\includegraphics[width=1.9cm]{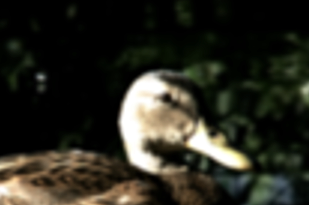} & 
\includegraphics[width=1.9cm]{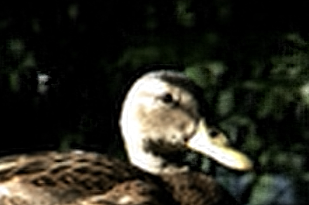}
&
\includegraphics[width=1.9cm]{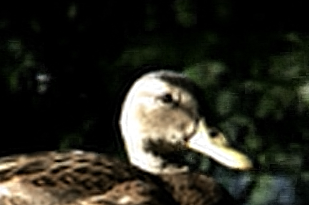}
&
\includegraphics[width=1.9cm]{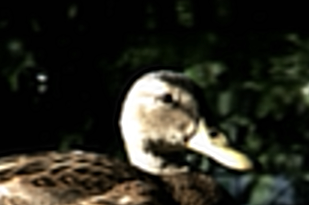}  
&
\includegraphics[width=1.9cm]{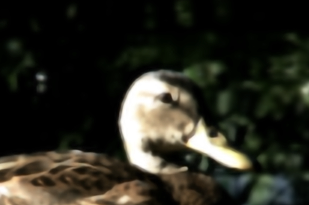} 
&
\includegraphics[width=1.9cm]{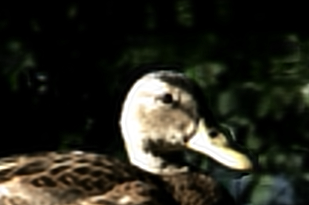} 
\\
\includegraphics[width=1.9cm]{duck_HR.png} & 
\includegraphics[width=1.9cm]{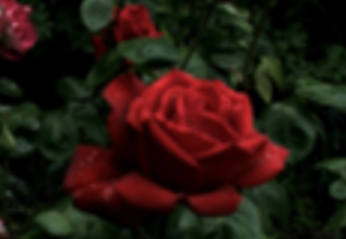} & 
\includegraphics[width=1.9cm]{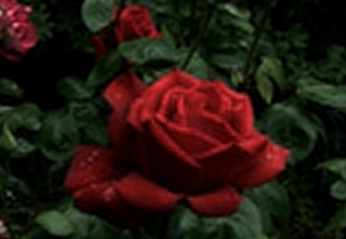}
&
\includegraphics[width=1.9cm]{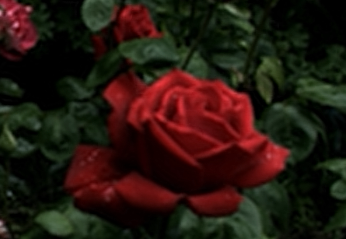}
&
\includegraphics[width=1.9cm]{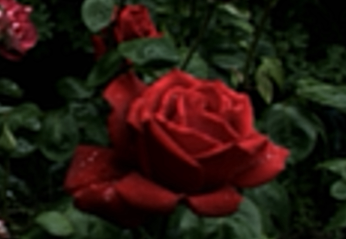}  
&
\includegraphics[width=1.9cm]{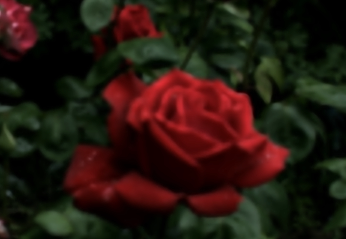} 
&
\includegraphics[width=1.9cm]{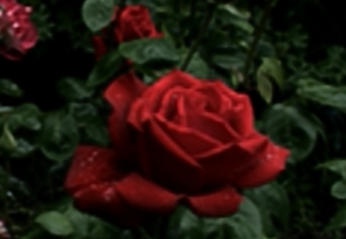} 
\\
\includegraphics[width=1.9cm]{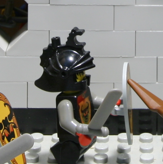} & 
\includegraphics[width=1.9cm]{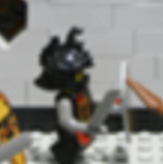} & 
\includegraphics[width=1.9cm]{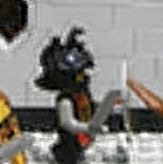}
&
\includegraphics[width=1.9cm]{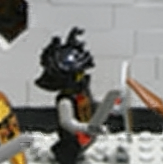}
&
\includegraphics[width=1.9cm]{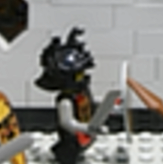}  
&
\includegraphics[width=1.9cm]{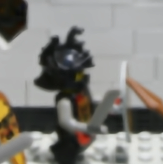} 
&
\includegraphics[width=1.9cm]{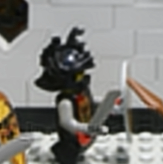} 
\\
\includegraphics[width=1.9cm]{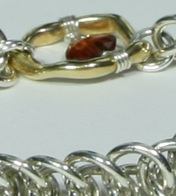} & 
\includegraphics[width=1.9cm]{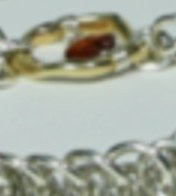} & 
\includegraphics[width=1.9cm]{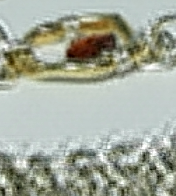}
&
\includegraphics[width=1.9cm]{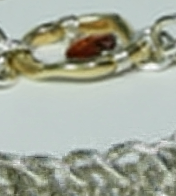}
&
\includegraphics[width=1.9cm]{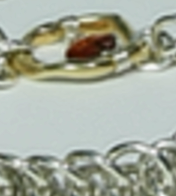}  
&
\includegraphics[width=1.9cm]{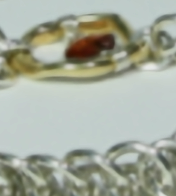} 
&
\includegraphics[width=1.9cm]{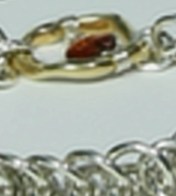} 
\\
\includegraphics[width=1.9cm]{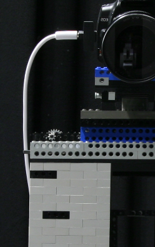} & 
\includegraphics[width=1.9cm]{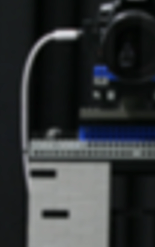} & 
\includegraphics[width=1.9cm]{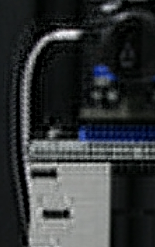}
&
\includegraphics[width=1.9cm]{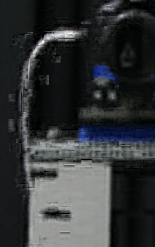}
&
\includegraphics[width=1.9cm]{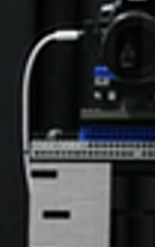}  
&
\includegraphics[width=1.9cm]{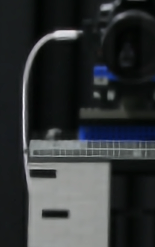} 
&
\includegraphics[width=1.9cm]{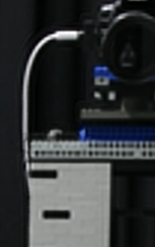} 
\end{tabular} 
\caption{Restored center view using different light field super-resolution algorithms. These are best viewed in color and by zooming on the views.}
\label{fig:subjective_Eval_results}
\end{figure*}

The results in table \ref{tbl:psnr_analysis_blurredx2} and table \ref{tbl:psnr_analysis_blurredx3} compare these super-resolution methods in terms of PSNR for magnification factors of $\times2$ and $\times3$ respectively.
Moreover, the results in Figure \ref{fig:subjective_Eval_results} show the centre views of light fields restored using these methods when considering a magnification factor of $\times3$.
The VDSR algorithm \cite{kim_2016} achieves on average a PSNR gain of 0.33 dB over bicubic interpolation.
One major limitation of VDSR is that it does not exploit the light field structure where each sub-aperture image is being restored independently.
The PCA+RR algorithm \cite{Farrugia2017} manages to restore more texture detail and is particularly effective to restore light fields with small disparities, which is the case of the INRIA light fields. This can be attributed to the fact that the PCA+RR does not compensate for disparities and therefore is not able to generalize to light fields containing disparities which were not considered during training, which is the case for the Stanford light fields.
In fact the Stanford light fields restored using PCA+RR contain blocking artefacts.
The BM+PCA+RR method \cite{Farrugia2017} extends this method by aligning the patch-volumes using block-matching.
This results in a more generalized method that achieves average PSNR gains of 2.52 dB and 1.76 dB over bi-cubic on the INRIA and Stanford light fields respectively.
Nevertheless, the light fields restored using this method may contain some artefacts especially near the edges.

\begin{table*}[tb]
\caption{PSNR using different light field super-resolution algorithms when considering a magnification factor of $\times 2$. For clarity  bold \textcolor{blue}{blue} marks the highest and bold \textcolor{red}{red} indicates the second highest score.}
\label{tbl:psnr_analysis_blurredx2}
\begin{center}
\begin{tabular}{|l|c|c|c|c|c|c|}
\hline
\bf{Light Field Name} & \bf{Bicubic} & \bf{PCA+RR \cite{Farrugia2017}} &  \bf{BM+PCA+RR \cite{Farrugia2017}} & \bf{LF-SRCNN \cite{Yoon2017}}& \bf{VDSR \cite{kim_2016}} & \bf{Proposed}\\ 
\hline
\hline
Bee 2 (INRIA) & 30.1673 &	\bf{\textcolor{blue}{34.1579}}	& 33.4655 & \bf{\textcolor{red}{33.8268}} & 30.4027 & 33.4915 \\
Dist. Church (INRIA) & 24.3059 &	\bf{\textcolor{red}{26.5071}} &	26.4571 & 25.9930 & 24.5419	& \bf{\textcolor{blue}{26.7502}}      \\
Duck (INRIA) & 23.5394 & \bf{\textcolor{blue}{26.7401}} &	26.2528	& 26.0713 &	23.8371 & \bf{\textcolor{red}{26.3777}} \\
Framed (INRIA) & 27.5974 &	\bf{\textcolor{blue}{30.7093}} &	30.4965 & 30.0697 &	27.8725 & \bf{\textcolor{red}{30.5365}} \\
Fruits (INRIA) & 28.4907 & 31.7884 &	\bf{\textcolor{red}{32.0002}} & 31.6820	& 28.7827 &	\bf{\textcolor{blue}{32.0789}}  \\
Mini (INRIA) & 27.6332 & \bf{\textcolor{blue}{30.4751}} &	30.0867 & 29.8941 & 27.9175 & \bf{\textcolor{red}{30.1601}} \\
Rose (INRIA) & 33.5436 & \bf{\textcolor{blue}{37.0791}} &	\bf{\textcolor{red}{36.9566}}	& 36.8245 &	33.7943 & 36.8416     \\
Amethyst (STANFORD) & 30.5227 &	\bf{\textcolor{blue}{32.5139}} &	\bf{\textcolor{red}{32.4262}}	& 32.2953 &	30.8360 & 32.2737 \\
Bracelet (STANFORD) & 26.4662 &	23.8183 &	28.2356 & \bf{\textcolor{red}{28.8858}} &	26.8523	& \bf{\textcolor{blue}{29.4046}} \\
Chess (STANFORD) & 30.2895 & 31.9292 &	\bf{\textcolor{red}{32.5708}} & 32.1922
& 30.6313 &	\bf{\textcolor{blue}{32.6123}} \\
Eucalyptus (STANFORD) & 30.7865 & 32.4162 & \bf{\textcolor{red}{32.4900}} & 32.1989  & 31.0431 &	\bf{\textcolor{blue}{32.6205}} \\
Lego Gantry (STANFORD) & 27.6235 &	28.0729	& 28.7230 & \bf{\textcolor{red}{29.8086}}	& 27.9998 & \bf{\textcolor{blue}{29.8112}} \\
Lego Knights (STANFORD) & 27.3794 &	27.8457 & \bf{\textcolor{red}{29.4664}} & \bf{\textcolor{blue}{29.5354}} & 27.7446 &	29.3177 \\
\hline
\hline
\end{tabular}
\end{center}
\end{table*}

\begin{table*}[tb]
\caption{PSNR using different light field super-resolution algorithms when considering a magnification factor of $\times 3$. For clarity  bold \textcolor{blue}{blue} marks the highest and bold \textcolor{red}{red} indicates the second highest score.}
\label{tbl:psnr_analysis_blurredx3}
\begin{center}
\begin{tabular}{|l|c|c|c|c|c|c|}
\hline
\bf{Light Field Name} & \bf{Bicubic} & \bf{PCA+RR \cite{Farrugia2017}} &  \bf{BM+PCA+RR \cite{Farrugia2017}} & \bf{LF-SRCNN \cite{Yoon2017}}& \bf{VDSR \cite{kim_2016}} & \bf{Proposed}\\ 
\hline
\hline
Bee 2 (INRIA) &  27.8623 &	31.2436	& 31.1886 & \bf{\textcolor{blue}{31.3945}}  & 28.2457 &	\bf{\textcolor{red}{31.2545}}
\\
Dist. Church (INRIA) & 23.3138 & \bf{\textcolor{blue}{24.7103}} & 24.6220	& 24.5874 &	23.7707 & \bf{\textcolor{red}{24.6535}} \\
Duck (INRIA) & 22.0702 & 23.9844 & 23.8083 & \bf{\textcolor{red}{24.0623}} &	22.5350 &	\bf{\textcolor{blue}{24.1549}} \\
Framed (INRIA) & 26.1627 & 28.0771 & \bf{\textcolor{red}{28.1587}} & 27.9157 & 26.8462 &	\bf{\textcolor{blue}{28.2954 }}\\
Fruits (INRIA) &  26.5269 & 29.0908 & 29.1969 & \bf{\textcolor{red}{29.2100}}	& 26.9021 &	\bf{\textcolor{blue}{29.5297}} \\
Mini (INRIA) & 26.3035 & \bf{\textcolor{blue}{28.4265}} &	28.3212 &	28.1731 &	26.7192 &	\bf{\textcolor{red}{28.4009}} \\
Rose (INRIA) &  31.7687	& \bf{\textcolor{red}{34.4692}} &	\bf{\textcolor{blue}{34.5754}} & 34.3064 &	32.0424 &	34.3392 \\
Amethyst (STANFORD) &  29.0665 & \bf{\textcolor{red}{31.0848}} &	\bf{\textcolor{blue}{31.1184}} & 30.5971 &	29.5618 & 30.4628 \\
Bracelet (STANFORD) &  24.1221 & 22.2654 &	25.3946 & \bf{\textcolor{red}{26.1013}}	&	24.3484 & \bf{\textcolor{blue}{26.2712}} \\
Chess (STANFORD) &  27.3679 & 29.5207 &	29.6773 & \bf{\textcolor{red}{30.0278}} &	27.6514 & \bf{\textcolor{blue}{30.1485}} \\
Eucalyptus (STANFORD) &  29.2136 &	\bf{\textcolor{red}{31.3459}} &	\bf{\textcolor{blue}{31.3547}} & 30.9433 &	29.4650 &	31.1772 \\
Lego Gantry (STANFORD) &  24.6721 &	25.3941 & 25.6383 & \bf{\textcolor{red}{26.8054}} &	24.7923 &	\bf{\textcolor{blue}{26.9466}} \\
Lego Knights (STANFORD) & 24.0182 &	24.4944 & 25.5768 & \bf{\textcolor{red}{26.0771}}	&	24.1256	& \bf{\textcolor{blue}{26.1358}} \\
\hline
\hline
\end{tabular}
\end{center}
\end{table*}

The LF-SRCNN method \cite{Yoon2017}, which uses deep learning to restore each sub-aperture image independently, was found to achieve a marginal gain over BM+PCA+RR (0.05 dB).
While, light fields restored using LF-SRCNN generally contain less artefacts compared to BM+PCA+RR, they risk to restore light fields which are angularly incoherent.
Nevertheless, our method achieves the best performance achieving an overall gain of 0.2 dB over the second-best performing algorithm LF-SRCNN.
This performance gain is more evident in the subjective results illustrated in Figure \ref{fig:subjective_Eval_results} where the restored light fields are sharper and are visually more pleasing.

As mentioned in section \ref{sec:ibp}, one problem with the proposed method is that the inverse warping is unable to perfectly restore the original disparities of the light field.
Nevertheless, the results in tables \ref{tbl:psnr_analysis_blurredx2}, \ref{tbl:psnr_analysis_blurredx3} and Figure \ref{fig:subjective_Eval_results} clearly show that our proposed method outperforms the other schemes even without the use of the iterative back-projection refinement strategy.

In order to fairly assess the contribution of iterative back-projection, we apply it as a post process for the two best performing methods, namely LF-SRCNN and our proposed scheme LR-LFSR. 
Tables \ref{tbl:psnr_analysis_blurredx2_ibp} and \ref{tbl:psnr_analysis_blurredx3_ibp} show the performance of the two best performing methods with and without iterative back projection as a post-processing step at magnification factors of $\times2$ and $\times3$ respectively.
It can be immediately noticed that IBP significantly improves the quality of both methods. 
Nevertheless, our method which uses iterative back projection as a post-processing step achieves the best performance achieving PSNR gains of 0.41 dB and 0.31 dB at magnification factors of $\times2$ and $\times3$ respectively over LF-SRCNN followed by iterative back projection.
It is important to notice that while the performance gain of our method over LF-SRCNN without IBP is around 0.23 dB and 0.12 dB at magnification factors of $\times2$ and $\times3$ respectively, this gain roughly doubles when both use IBP as a post processing step. 
This indicates that since LF-SRCNN processes each view independently, IBP only corrects inconsistencies between the low-resolution and restored light fields. Apart from this distortion, LR-LFSR-IBP corrects the distortions caused by the inverse warping process which provides light fields which are more visually pleasing and with smoother transitions across views.
Supplementary multimedia files uploaded on ScholarOne show some sample restored light field.
Supplementary material is available on the project's website\footnote{\url{https://goo.gl/8DDsDi}} while the code of the LR-LFSR will be made available upon publication.

\begin{table*}[tb]
\caption{PSNR obtained with the best two methods at a magnification of $\times2$ with and without iterative back projection as a post process. }
\label{tbl:psnr_analysis_blurredx2_ibp}
\begin{center}
\begin{tabular}{|l|c|c|c|c|c|}
\hline
\bf{Light Field Name} & \bf{Bicubic} & \bf{LF-SRCNN} &  \bf{LF-SRCNN-IBP} & \bf{LR-LFSR}& \bf{LR-LFSR-IBP}\\ 
\hline
\hline
Bee 2 (INRIA) & 30.1673 &  33.8268 & 34.7257 & 33.4915 & \bf{35.0306}\\
Dist. Church (INRIA) & 24.3059 & 25.9930 & 26.7415 & 26.7502 & \bf{27.1527} \\
Duck (INRIA) & 23.5394 & 26.0713 & 27.293 & 26.3777 & \bf{27.6928} \\
Framed (INRIA) & 27.5974 & 30.0697 & 31.0592 & 30.5365 & \bf{31.7185}\\
Fruits (INRIA) &  28.4907 & 31.6820 & 32.6581 & 32.0789 & \bf{33.3370} \\
Mini (INRIA) & 27.6332 & 29.8941 & 30.5193 & 30.1601 & \bf{30.9899} \\
Rose (INRIA) &  33.5436 & 36.8245 & 37.4991 & 36.8416 & \bf{37.8970} \\
Amethyst (STANFORD) &  30.5227 & 32.2953 & 33.5662 & 32.2737 & \bf{33.8686} \\
Bracelet (STANFORD) &  26.4662 & 28.8858 & 30.6367 & 29.4046 & \bf{30.9579} \\
Chess (STANFORD) &  30.2895 & 32.1922 & 33.6715 & 32.6123 & \bf{34.1438} \\
Eucalyptus (STANFORD) &  30.7865 & 32.1989 & 33.0700 & 32.6205 & \bf{33.4550}\\
Lego Gantry (STANFORD) & 27.6235 & 29.8086 & 31.5071 & 29.8112 & \bf{31.8631} \\
Lego Knights (STANFORD) & 27.3794 & 29.5354 & 31.2148 & 29.3177 & \bf{31.3240} \\
\hline
\hline
\end{tabular}
\end{center}
\end{table*}

\begin{table*}[tb]
\caption{PSNR obtained with the best two methods at a magnification of $\times3$ with and without iterative back projection as a post process. }
\label{tbl:psnr_analysis_blurredx3_ibp}
\begin{center}
\begin{tabular}{|l|c|c|c|c|c|}
\hline
\bf{Light Field Name} & \bf{Bicubic} & \bf{LF-SRCNN} &  \bf{LF-SRCNN-IBP} & \bf{LR-LFSR}& \bf{LR-LFSR-IBP}\\ 
\hline
\hline
Bee 2 (INRIA) & 27.8623 & 31.3945 & 32.4662 & 31.2545 & \bf{32.7661}\\
Dist. Church (INRIA)  & 23.3138 & 24.5874 & \bf{25.2440} & 24.6535 & 25.1669\\
Duck (INRIA)  & 22.0702 & 24.0623 & 24.7698 & 24.1549 & \bf{25.0554}\\
Framed (INRIA) & 26.1627 & 27.9157 & 28.9111 & 28.2954 & \bf{29.5017}\\
Fruits (INRIA)  & 26.5269 & 29.2100 & 30.1651 & 29.5297 & \bf{30.7844}\\
Mini (INRIA)  & 26.3035 & 28.1731 & 28.6895 & 28.4009 & \bf{29.1335}\\
Rose (INRIA)  & 31.7687 & 34.3064 & 34.9356 & 34.3392 & \bf{35.4656}\\
Amethyst (STANFORD)  & 29.0665 & 30.5971 & 31.5991 & 30.4628 & \bf{31.8143}\\
Bracelet (STANFORD)  & 24.1221 & 26.1013 & 27.0010 & 26.2712 & \bf{27.3423}\\
Chess (STANFORD) & 27.3679 & 30.0279 & 31.1502 & 30.1485 & \bf{31.4388} \\
Eucalyptus (STANFORD) & 29.2136 & 30.9433 & 31.5263 & 31.1772 & \bf{31.9161}\\
Lego Gantry (STANFORD)  & 24.6721 & 26.8054 & 27.8395 & 26.9466 & \bf{27.8852}\\
Lego Knights (STANFORD) & 24.0182& 26.0771 & 27.6550 & 26.1358 & \bf{27.7168}\\
\hline
\hline
\end{tabular}
\end{center}
\end{table*}

\bibliographystyle{IEEEtran}
\bibliography{IEEEabrv,lfsr}

\section{Conclusion}
\label{sec:conclusion}
In this paper, we have proposed a novel spatial light field super-resolution algorithm able to reconstruct high quality coherent light fields.
We have shown that the information in a light field can be efficiently compacted by aligning the sub-aperture images using optical flow followed by low-rank matrix approximation. The low rank approximation of the aligned light field gives an embedding in a lower dimensional space which is super-resolved using deep learning. All aligned views of the high-resolution light field can be reconstructed from the super-resolved embedding by simple linear combinations.
These views are then inverse warped to restore the disparities of the original light field. Holes corresponding to dis-occlusions or cracks resulting from the inverse warping are filled in using a novel diffusion based inpainting algorithm which diffuses known pixels in the EPI along dominant orientations computed in the low-resolution EPI.

Extensive simulations show that the proposed method manages to generalize well, i.e. manages to successfully restore light fields whose disparities are considerably different from those used during training.
These results also show that our proposed method is competitive and most of the time superior to existing state-of-the-art light field super-resolution algorithms, including a recent approach which adopts deep learning to restore each sub-aperture image independently.
One major limitation of the proposed scheme is that the inverse warping process is not able to restore the original disparities and produces some distortion caused by rounding errors. 
We proposed here to use the classical iterative back-projection as a post processing step.
Simulation results clearly show the benefit of using IBP as a post processing of the super-resolved light field and demonstrate that the proposed method with IBP achieves the best performance, outperforming LF-SRCNN followed by IBP by 0.4 dB. 


%

\ifCLASSOPTIONcompsoc
  \section*{Acknowledgments}
\else
  \section*{Acknowledgment}
\fi

This project has been supported in part by the EU H2020 Research and Innovation 
Programme under grant agreement No 694122 (ERC advanced grant CLIM).

\ifCLASSOPTIONcaptionsoff
  \newpage
\fi

\begin{IEEEbiography}[{\includegraphics[width=1in,height=1.25in,clip,keepaspectratio]{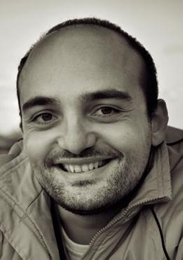}}]{Reuben A. Farrugia}
(S’04, M’09, SM’17) received the first
degree in Electrical Engineering from the University of
Malta, Malta, in 2004, and the Ph.D. degree from the
University of Malta, Malta, in 2009.
In January 2008 he was appointed Assistant Lecturer with the same department
and is now a Senior Lecturer. 
He has been in technical and organizational committees of several national and international conferences. In particular, he served as General-Chair on the IEEE Int. Workshop on Biometrics and Forensics (IWBF) and as Technical Programme Co-Chair on the IEEE Visual Communications and Image Processing (VCIP) in 2014.  He has been contributing as a reviewer of several journals and conferences, including IEEE Transactions on Image Processing, IEEE Transactions on Circuits and Systems for Video and Technology and IEEE Transactions on Multimedia. On September 2013 he was appointed as National Contact Point of the European Association of Biometrics (EAB).
\end{IEEEbiography}

\begin{IEEEbiography}[{\includegraphics[width=1in,height=1.25in,clip,keepaspectratio]{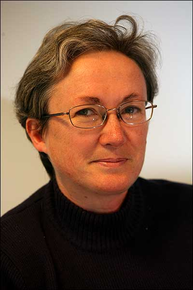}}] {Christine Guillemot} IEEE fellow, is “Director of Research” at INRIA, head of a research team
dealing with image and video modeling, processing, coding and communication. She holds a Ph.D. degree from ENST
(Ecole Nationale Superieure des Telecommunications) Paris, and an “Habilitation for Research Direction” from the
University of Rennes. From 1985 to Oct. 1997, she has been with FRANCE TELECOM, where she has been involved
in various projects in the area of image and video coding for TV, HDTV and multimedia. From Jan. 1990 to mid
1991, she has worked at Bellcore, NJ, USA, as a visiting scientist. She has (co)-authored 24 patents, 9 book chapters,
60 journal papers and 140 conference papers. She has served as associated editor (AE) for the IEEE Trans. on Image
processing (2000-2003), for IEEE Trans. on Circuits and Systems for Video Technology (2004-2006) and for IEEE Trans. On Signal Processing
(2007-2009). She is currently AE for the Eurasip journal on image communication, IEEE Trans. on Image Processing (2014-2016) and member of the editorial board for the IEEE Journal on
selected topics in signal processing (2013-2015). She is a member of the IEEE IVMSP technical committee.
\end{IEEEbiography}


\end{document}